\documentclass[default,iicol]{sn-jnl}

\usepackage{graphicx}%
\usepackage{multirow}%
\usepackage{amsmath,amssymb,amsfonts}%
\usepackage{amsthm}%
\usepackage{mathrsfs}%
\usepackage[title]{appendix}%
\usepackage{xcolor}%
\usepackage{textcomp}%
\usepackage{manyfoot}%
\usepackage{booktabs}%
\usepackage{algorithm}%
\usepackage{algorithmicx}%
\usepackage{algpseudocode}%
\usepackage{listings}%

\usepackage{array}
\usepackage{soul} 
\usepackage{pifont}
\usepackage{tablefootnote}
\usepackage{tabularx}
\usepackage{soul}
\newcolumntype{Y}{>{\centering\arraybackslash}X}

\theoremstyle{thmstyleone}%
\theoremstyle{thmstyletwo}%

\theoremstyle{thmstylethree}%

\newcommand{\dataset}{\texttt{MMNERD}}
\newcommand{\model}{\texttt{2M-NER}}

\begin{document}

\title[2M-NER: Contrastive Learning for Multilingual and Multimodal NER with Language and Modal Fusion]{2M-NER: Contrastive Learning for Multilingual and Multimodal NER with Language and Modal Fusion}

\author[1]{\fnm{Dongsheng} \sur{Wang}}\email{wangdsh@cupl.edu.cn}
\author[2]{\fnm{Xiaoqin} \sur{Feng}}\email{xiaoqin.feng@mobvoi.com}
\author*[3]{\fnm{Zeming} \sur{Liu}}\email{zmliu@buaa.edu.cn}
\author[4]{\fnm{Chuan} \sur{Wang}}\email{wangchuan@iie.ac.cn}

\affil[1]{\orgdiv{The Department of Science and Technology Teaching}, \orgname{China University of Political Science and Law}, \orgaddress{\city{Beijing}, \postcode{102249},\country{China}}}
\affil[2]{\orgname{Mobvoi AI Lab}, \orgaddress{\city{Beijing}, \postcode{100044},\country{China}}}
\affil*[3]{\orgdiv{School of Computer Science and Engineering}, \orgname{Beihang University}, \orgaddress{\city{Beijing}, \postcode{100191},\country{China}}}
\affil[4]{\orgdiv{Institute of Information Engineering}, \orgname{CAS}, \orgaddress{\city{Beijing}, \postcode{100085},\country{China}}}

\abstract{Named entity recognition (NER) is a fundamental task in natural language processing that involves identifying and classifying entities in sentences into pre-defined types. It plays a crucial role in various research fields, including entity linking, question answering, and online product recommendation. Recent studies have shown that incorporating multilingual and multimodal datasets can enhance the effectiveness of NER. This is due to language transfer learning and the presence of shared implicit features across different modalities. However, the lack of a dataset that combines multilingualism and multimodality has hindered research exploring the combination of these two aspects, as multimodality can help NER in multiple languages simultaneously. In this paper, we aim to address a more challenging task: multilingual and multimodal named entity recognition (MMNER), considering its potential value and influence. Specifically, we construct a large-scale MMNER dataset with four languages (English, French, German and Spanish) and two modalities (text and image). To tackle this challenging MMNER task on the dataset, we introduce a new model called \model{}, which aligns the text and image representations using contrastive learning and integrates a multimodal collaboration module to effectively depict the interactions between the two modalities. Extensive experimental results demonstrate that our model achieves the highest F1 score in multilingual and multimodal NER tasks compared to some comparative and representative baselines. Additionally, in a challenging analysis, we discovered that sentence-level alignment interferes a lot with NER models, indicating the higher level of difficulty in our dataset.}




\maketitle

\section{Introduction}

Named entity recognition (NER) is one of the basic tasks in natural language processing which aims to locate and classify specific things into pre-defined types, such as diseases, products, monetary values, etc. It is an important contributor to different research fields, including question answering \cite{DBLP:journals/isci/CuiPXHHL23, DBLP:journals/isci/DuJYY23}, automatic text-video retrieval \cite{DBLP:conf/cvpr/Li0LNH22} and machine translation \cite{DBLP:conf/dasfaa/YangYMYGHZZLW23, DBLP:conf/eacl/GuerreiroVM23}, as it enhances the understanding and interpretation of textual information in these tasks. Early studies recognize entities with long short-term memory (LSTM), attention-based technique, or convolution neural network (CNN) \cite{DBLP:journals/corr/HuangXY15, DBLP:conf/acl/MaH16}. owever, recent research has demonstrated that Transformer-based structures surpass these earlier methods in NER tasks \cite{DBLP:conf/acl/YuJYX20, DBLP:conf/aaai/ZhangWLWZZ21}. Despite these advancements, most of early studies solely utilized monolingual and unimodal text to recognize entities, which is insufficient \cite{DBLP:journals/tkde/LiCFW22, DBLP:conf/aaai/Agarwal22, DBLP:conf/acl/0001WTXXH0Z22}. For example, text-only methods ignore the helpfulness of corresponding images, leading to lower accuracy in entity recognition.

Some other researchers leverage multilingual and multimodal information to solve the aforementioned issues \cite{DBLP:conf/emnlp/SchmidtVG22a, DBLP:conf/wsdm/ZhangYLL23, DBLP:conf/eacl/KulkarniPRWWXY23}. On the one hand, some studies on multilingual NER \cite{DBLP:conf/acl/ZhangMCXZ21, DBLP:conf/semeval/BorosGMD22} have found that knowledge transfer from one language to another language is useful for zero-resource NER and cross-lingual NER. Hence, they have designed models to take advantage of multilingual features of multiple languages.
On the other hand, some works on multimodal NER \cite{DBLP:conf/aaai/0001FLH18, DBLP:conf/sigir/ChenZLDTXHSC22, DBLP:conf/naacl/WangGJJBWHT22} have combined the textual modality with the visual modality so that both explicit and implicit visual information can be exploited to help improve the performance of their models.
However, these studies have a limitation that they often focus on  multilingual or multimodal NER, with  many public datasets availiable for multilingual NER \cite{DBLP:conf/conll/Sang02, DBLP:conf/conll/SangM03, DBLP:conf/acl/PanZMNKJ17} and multimodal NER \cite{DBLP:conf/aaai/0001FLH18, DBLP:conf/acl/JiZCLN18, DBLP:conf/acl/SuiT00020} respectively. However, in the real world, multilingual and multimodal coexist, making it valuable for models to utilize the multilingual features of the multilingual text and integrate the related visual information with the textual information. Thus, it is highly beneficial to develop a multilingual and multimodal NER dataset to facilitate future research in this area.

To address the above issue and advance MMNER, we make the following attempts and efforts. We firstly build a large human-annotated \textbf{M}ultilingual and \textbf{M}ultimodal \textbf{NER} \textbf{D}ataset (\dataset{}), which is constructed from a large multilingual dataset and an English multimodal dataset via transformation, translation and human annotation. 
Concretely, we mark all appearances of entities belonging to four categories (person, location, organization, and miscellaneous) in a widely used multilingual dataset called mBART50\footnote{https://huggingface.co/datasets/flax-community/conceptual-12m-mbart-50-multilingual/tree/main} and a famous multimodal dataset Twitter-2017 \cite{DBLP:conf/acl/JiZCLN18}. As a result, \dataset{} has become the first public dataset which supports both multilingual and multimodal.
Its total number of sentences across four languages (English, French, Spanish and German) is 42,908 and the in-depth statistical analysis of \dataset{} is shown in Table \ref{MMNER_statistics}. Besides, the detailed comparison between \dataset{} and previous NER datasets in \ref{sec:dataset_comparison} highlights the multilingual and multimodal characteristics of our dataset.

\begin{figure*}[t]
\centering
\includegraphics[scale=0.40]{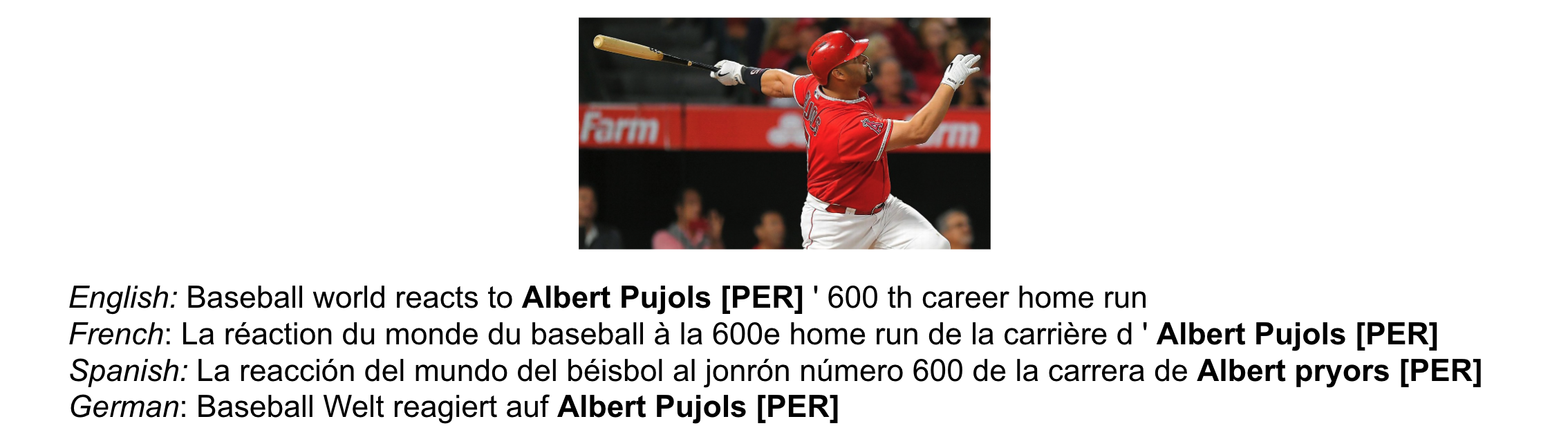}
\caption{An instance of multilingual and multimodal NER. The types of all entities are enclosed in brackets.}
\label{example1}
\end{figure*}

Based on \dataset{} and its new MMNER task, we propose a novel \textbf{M}ultilingual and \textbf{M}ultimodal \textbf{NER} model named \model{}. Concretely, considering the different architectures of various image encoders, \model{} leverages ViT \cite{DBLP:conf/iclr/DosovitskiyB0WZ21} and ResNet \cite{DBLP:conf/cvpr/HeZRS16} to extract patch features and convolution features, respectively. Meanwhile, instead of directly blending textual and visual modalities in many previous works, contrastive learning is applied in \model{} for modal alignment, which is implemented by the contrastive loss in the multimodal alignment module. 
For instance, in Fig. \ref{example1}, the entity \textit{Albert Pujols} in the English sentence aligns with the area of a baseball player within the image. In the representation space, the entity \textit{Albert Pujols} and its corresponding area should have similar embeddings, so that they will have bigger similarity score. Besides, \model{} utilizes two cross-attention Transformer \cite{DBLP:journals/apin/SunZTXW24} layers to build modal interactions between text and images.

Moreover, we conduct extensive experiments on \dataset{}. A set of classical and competitive NER methods are selected for comparison. Specifically, we first utilize some representative text-based NER models such as BiLSTM-CRF \cite{DBLP:journals/corr/HuangXY15}, HBiLSTM-CRF {\cite{DBLP:conf/naacl/LampleBSKD16}, BERT \cite{DBLP:conf/naacl/DevlinCLT19}, etc. After that, to illustrate the help of visual representation to entity recognition, we pick out several representative multimodal NER models, such as UMT \cite{DBLP:conf/acl/YuJYX20}, UMGF \cite{DBLP:conf/aaai/ZhangWLWZZ21}, and MKGFormer \cite{DBLP:conf/sigir/ChenZLDTXHSC22}, on our dataset. In terms of linguistic aspect, all the models we use are tested on all four languages and also each of them. These experiments prove the effectiveness of \model{} in the new MMNER task.

Our work can be primarily characterized by the following major contributions:

\begin{itemize}
\item[(1)] We construct \dataset{}, a large-scale human-annotated MMNER dataset, which has four languages and two modalities. So far as we know, this is the first public dataset that supports both multilingual and multimodal NER.
\item[(2)] To facilitate research on this dataset, our paper introduces a novel model called \model{} for MMNER, which leverages contrastive learning to align the language and vision representations.
\item[(3)] Based on \model{}, comparative experiments illustrate the efficiency of our methodology, while more relevant experiments indicate that the assistance of multilingual and visual modal can enhance the performance of MMNER.
\end{itemize}

The following sections give the organization of this paper. In Section \ref{sec:related_work}, we summarize the relevant literature on multilingual NER and multimodal NER. Section \ref{sec:dataset_construction} describes the procedure of dataset construction in detail. A summary of our proposed model and the illustration of each component are presented in Section \ref{sec:methodology}. The comparative experiments and ablation study conducted on \dataset{} are presented in Section \ref{sec:experiments}. To summarize our work, Section \ref{sec:conclusion} provides a comprehensive overview and lists some future work.

\section{Related Work}\label{sec:related_work}

In this section, we review two groups of relevant studies closely associated with MMNER: multilingual named entity recognition and multimodal named entity recognition.

\subsection{Multilingual Named Entity Recognition}

Named entity recognition (NER) is a general task that provides foundational support for a range of natural language processing (NLP) tasks. As the research on multilingual models deepens, monolingual NER gradually migrates to multilingual NER, which provides a solid foundation for multilingual NLP tasks such as question answering \cite{DBLP:conf/coling/SenAS22, DBLP:conf/www/PerevalovBDN22}, information retrieval \cite{DBLP:conf/cikm/WangZZGW021, DBLP:conf/emnlp/SunD20}, relation extraction \cite{DBLP:conf/acl/BhartiyaBM22, DBLP:conf/acl/RathoreS22}, etc. In previous studies, NER relied on statistical modeling of annotated data, which is extremely arduous and expensive. For example, Joel et al. \cite{DBLP:journals/ai/NothmanRRMC13} proposed a data annotations method by utilizing the textual content and structural framework of Wikipedia. This enormous, accessible, and multilingual data helps the model to surpass domain-specific models in terms of performance. Recently, many researchers have presented various methods to obtain the most advanced outcomes in NER. Specifically, Shervin et al. \cite{DBLP:conf/coling/MalmasiFFKR22} presented MultiCoNER, a vast multilingual corpus for NER that encompasses three domains and eleven different languages, including sub-corpora featuring multilingual and code-mixed content. They applied two NER models to showcase the difficulty and validity of the dataset. Emanuela et al. \cite{DBLP:conf/semeval/BorosGMD22}, through their analysis of the MultiCoNER dataset, discovered that incorporating supplementary contextual information from the training data can enhance the performance of NER on shorter text samples. Additionally, Shervin et al. \cite{DBLP:conf/semeval/MalmasiFFKR22} fused external knowledge into transformer models to achieve the best performance. Moreover, CoNLL2002 \cite{DBLP:conf/conll/Sang02}, CoNLL2003 \cite{DBLP:conf/conll/SangM03}, and WikiAnn \cite{DBLP:conf/acl/PanZMNKJ17} are also commonly used multilingual NER datasets. In comparison to them, the dataset we constructed incorporates multimodal characteristics.

Owing to the advancements in pre-trained language models (PLM), multilingual NER has started to utilize these models as embeddings to improve its performance \cite{DBLP:conf/acl-bsnlp/EmelyanovA19, DBLP:conf/acl-bsnlp/ArkhipovTKS19}. Genta et al. \cite{DBLP:conf/rep4nlp/WinataLF19} and Wu et al. \cite{DBLP:conf/aaai/WuLWCKHL20} proposed the different meta-embedding methods to represent multilingual text. In the meantime, several researchers have employed multilingual models that support over one hundred languages as a foundation for transfer learning to other language models, which enables them to achieve superior performance compared to many neural methods and multilingual pre-trained models \cite{DBLP:conf/acl-bsnlp/ArkhipovTKS19, DBLP:conf/emnlp/SchmidtVG22a}. In summary, many multilingual methods for NER have been proposed and validated in rare languages and industrial areas, but further in-depth research is still required. Different from these methods, our approach incorporates additional visual information to assist in multilingual NER.

\subsection{Multimodal Named Entity Recognition}

Multimodal NER, similar to multilingual NER, has recently drawn the attention of academics due to the abundance of user-generated graphic and textual data on social media platforms like Twitter. Base on Twitter data, researchers have built two widely used datasets, Twitter-2015 \cite{DBLP:conf/aaai/0001FLH18} and Twitter-2017 \cite{DBLP:conf/acl/JiZCLN18}, which are text-image corpora. However, our dataset stands out in terms of its larger scale and multilingual nature. Through employing these multimodal NER datasets, numerous studies have leveraged the corresponding images to help recognize entities within the text. During the initial phases of studies, Zhang et al. \cite{DBLP:conf/aaai/0001FLH18}, Lu et al. \cite{DBLP:conf/acl/JiZCLN18}, and Moon et al. \cite{DBLP:conf/naacl/MoonNC18} adopted long short-term memory (LSTM) networks to obtain text representations and convolution neural network (CNN) to obtain features from images.
The methods used for text and image feature extraction and fusion in these works are coarse-grained and simple. Recently, pre-trained models like BERT have been employed in multimodal NER to generate better text representations. Specifically, Yu et al. \cite{DBLP:conf/acl/YuJYX20} employed a BERT encoder to obtain contextualized representations for input sentences. They also proposed a multimodal interaction module that generates word representations with knowledge of images, and visual representations with knowledge of words. Similarly, Zhao et al. \cite{DBLP:conf/mm/ZhaoLWXD22} used the pre-trained model BERT for text encoding, and they proposed a relation-based graph convolutional architecture to examine the extrinsic matching relationships among pairs of texts and images.
To establish a text-image relation, Sun et al. \cite{DBLP:conf/aaai/0006W0SW21} built a propagation-based BERT model for multimodal NER by integrating soft or hard gates.

In the meantime, several researchers have found that it was useful for multimodal NER to find fine-grained visual objects and filter out irrelevant visual objects with an object detection model. To this end, Zheng et al. \cite{DBLP:journals/tmm/ZhengWWCL21} built an adversarial attention-based network with Mask RCNN \cite{DBLP:journals/apin/LiK23} to extract visual object features, and constructed a common subspace for text and image modalities by adversarial learning. Similarly, Wu et al. \cite{DBLP:conf/mm/WuZCCL020} introduced a neural network that employs a pre-trained Mask RCNN network to recognize objects and utilizes a dense co-attention mechanism to learn the relationships between visual objects and textual entities.
To exploit semantic relevance between different modalities, Zhang et al. \cite{DBLP:conf/aaai/ZhangWLWZZ21} designed a multimodal graph that is constructed from words and visual objects. Besides, Wang et al. \cite{DBLP:conf/naacl/WangGJJBWHT22} applied optical character recognition (OCR) techniques to obtain object tags from images and mixed text and image modalities via transformer-based embeddings. The above methods require pre-trained object detectors. To reduce the dependence on them, Chen et al. \cite{DBLP:conf/sigir/ChenZLDTXHSC22} introduced a unified model for multimodal knowledge graph completion tasks, which adopts ViT \cite{DBLP:conf/iclr/DosovitskiyB0WZ21} for image embedding. 
Similarly, our model also does not rely on pre-trained object detectors. Moreover, we align text and visual representations before integrating multimodal information, which is effective and distinguishes our model from previous approaches.

\section{Dataset Construction}\label{sec:dataset_construction}

Although building a multilingual and multimodal NER dataset can greatly facilitate the development of the MMNER task, no research has yet undertaken this endeavor. Hence, we plan to build one. In this part, we will first introduce our complete process of dataset annotation. Afterwards, we will present the statistics of our MMNER dataset. Finally, we will compare the MMNER dataset with other NER datasets.

\subsection{Dataset Annotation}

During the data annotation phase, we first select and process the data. Then, some data commissioners are trained to improve the quality of annotations. Finally, these commissioners spend three months completing all the annotation work. Obviously, it is easy to find a standalone multilingual or multimodal dataset. Therefore, we can transform a multilingual dataset to a MMNER dataset, or translate a single-language multimodal dataset into a MMNER dataset. The former is more reliable as it reduces the burden of translation. After conducting thorough investigations and careful consideration, we have opted for a combination approach. \textit{Firstly}, we choose a large multilingual dataset called mBART50, which is released by Hugging Face\footnote{https://huggingface.co} and contains 2.5 million pairs of images and texts in four languages (English, French, German, and Spanish), translated via mBART-50 \cite{DBLP:journals/tacl/LiuGGLEGLZ20}. In this dataset, each image is associated with its respective URL. Through downloading the images using these URLs, we get a basic multilingual and multimodal dataset. Besides, considering the cost of NER annotation, we choose more English data. Concretely, the selected number of image-text pairs in English, French, German and Spanish is 30000, 5000, 5000, and 5000 respectively. To ensure the quality and efficiency of dataset annotation, we firstly utilize spaCy\footnote{https://spacy.io/} to recognize entities and remove image-text pairs whose entities number is less than two. All entities, except person, location, and organization, are marked as \textit{MISC}. \textit{Secondly}, note that person entities in mBART50 are replaced with the token \emph{<PERSON>}. Therefore, Twitter-2017 \cite{DBLP:conf/acl/JiZCLN18} is also added into our MMNER dataset to complement person entity annotations. Moreover, since Twitter-2017 is an English multimodal dataset, we have translated it into French, German, and Spanish with the help of Baidu Translate\footnote{https://fanyi-api.baidu.com}. Specifically, 2982 twitters have been translated into these three languages.

\begin{table*}[htbp]
\renewcommand\arraystretch{1.5}
\caption{Examples of \dataset{}. Note that the first text-image pair comes from Twitter-2017, and the other text-image pairs come from mBART50.}
\label{data_examples}
\centering
\begin{tabular*}{\textwidth}{@{\extracolsep{\fill}}  m{0.66\textwidth}  >{\centering\arraybackslash}m{0.34\textwidth} }
  \toprule
  Text & Image \\
  \midrule
  English: Baseball world reacts to \textbf{Albert Pujols} \textbf{[PER]} ' 600 th career home run \par 
  French: La réaction du monde du baseball à la 600e home run de la carrière d ' \textbf{Albert Pujols} \textbf{[PER]} \par 
  Spanish: La reacción del mundo del béisbol al jonrón número 600 de la carrera de \textbf{Albert pryors} \textbf{[PER]} \par 
  German: Baseball Welt reagiert auf \textbf{Albert Pujols} \textbf{[PER]} &
  \includegraphics[width=4cm]{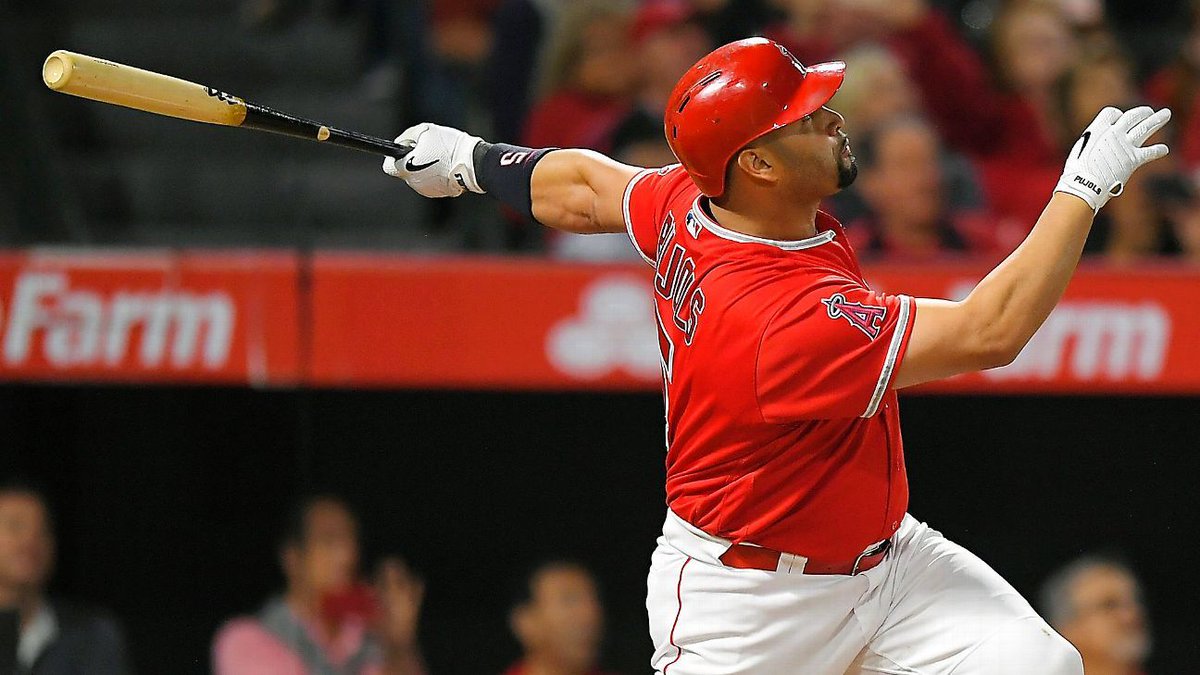} \\
  \midrule
  English: An elephant observes zebras and other animals in \textbf{Namibia} \textbf{[LOC]} ' s \textbf{Etosha National Park} \textbf{[LOC]} \par 
  French: Un éléphant observe des zèbres et d'autres animaux dans le \textbf{parc national d'Etosha} \textbf{[LOC]} en \textbf{Namibie} \textbf{[LOC]} \par 
  Spanish: Un elefante observa cebras y otros animales en el \textbf{Parque nacional Etosha} \textbf{[LOC]} de \textbf{Namibia} \textbf{[LOC]} \par 
  German: Ein Elefant beobachtet Zebras und andere Tiere im \textbf{Etosha Nationalpark} \textbf{[LOC]} \textbf{Namibias} \textbf{[LOC]} \par & 
  \includegraphics[width=4cm]{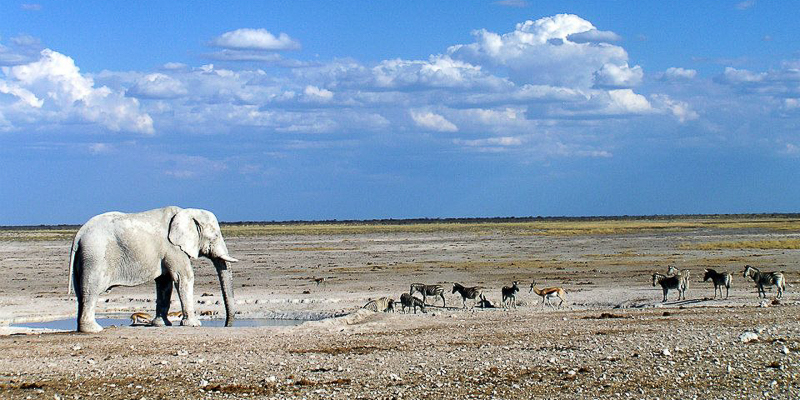} \\
  \midrule
  English: Fireworks Over The \textbf{Thames} \textbf{[LOC]} For \textbf{New Year} \textbf{[MISC]} \par 
  French: Feux d'artifice sur la \textbf{Tamise} \textbf{[LOC]} pour le \textbf{nouvel an} \textbf{[MISC]} \par 
  Spanish: Fuegos artificiales en el \textbf{Támesis} \textbf{[LOC]} de \textbf{año nuevo} \textbf{[MISC]} \par 
  German: Feuerwerk über der \textbf{Themse} \textbf{[LOC]} für \textbf{Neujahr} \textbf{[MISC]} & 
  \includegraphics[width=4cm]{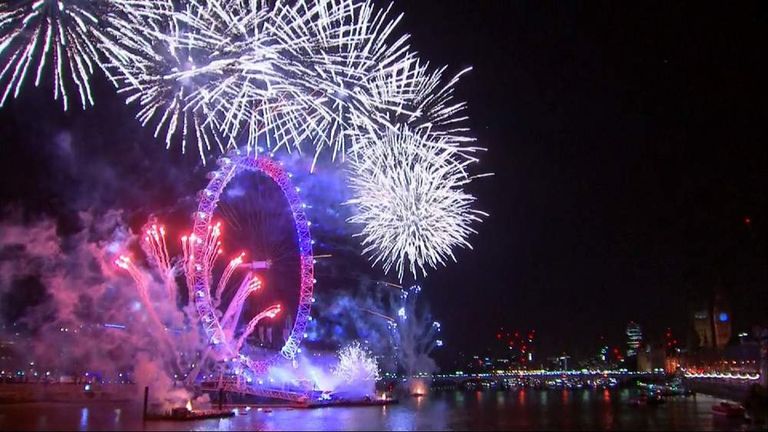} \\
  \bottomrule
\end{tabular*}
\end{table*}

Before annotating the MMNER dataset, all data commissioners are initially trained on a subset corpus multiple times in advance, aiming to ensure consistency and reliability in intra- and inter-rater annotation. The accuracy of dataset annotation among all annotators improved from 81\% to 90\% after the training process.

During the data annotation process, 13 annotators help us to annotate the MMNER data manually with a detailed annotation instruction and the gold examples preprocessed by spaCy. Concretely, for each sentence, we ask for two annotators to perform annotation and one inspector for checking. Following the evaluation method in previous work \cite{DBLP:conf/acl/SuiT00020}, we evaluate the reliability between annotators via Cohen's kappa coefficient \cite{kappa_coefficient} and its value is 0.96, which indicates the data quality is relatively high. In the end, the processed mBART50 and Twitter-2017 data are combined and partitioned into training, validation, and test sets. Some examples of our MMNER dataset are listed in Table \ref{data_examples}.
In those examples, the first image from Twitter-2017 depicts a distinct person entity which is the core part of the corresponding sentences. The other images come from mBART50, and each of them corresponds to some entities and many other words in those sentences, which makes the MMNER task on mBART50 is extremely challenging.

\subsection{Dataset Statistics and Evaluation}

The details of the text in our MMNER dataset are summarized in Table \ref{MMNER_statistics}. The proportion of training, development and test sets is 8:1:1. We have counted the number of 4 classes of entities (person, location, organization and miscellaneous) for each language. Additionally, we have also conducted statistics on the number of samples in the training, validation, and test sets, as well as a summary for each entity type. The total number of entities for the four classes is 89,019.

Each sentence has a corresponding image, and we store all 33,965 images in a dedicated folder. Besides, a default image\footnote{An image named 17\_06\_4705.jpg with the words "image not found".} is added to replace a few broken images in Twitter-2017. The MMNER dataset is released for further studies via this link: \href{https://github.com/wangdsh/MMNERD}{https://github.com/wangdsh/MMNERD}.

After the data annotation, we randomly selected 300 sentences to test the quality of the dataset. By manual review of each entity annotation, we find that 626 out of 631 entities are correctly annotated, indicating that the dataset's quality is well guaranteed.

\begin{table*}[htbp]
\caption{Statistics of the MMNER dataset.}
\label{MMNER_statistics}
\begin{tabularx}{\textwidth}{@{} YYYYYYYYYYYYYl @{}}
  \toprule
   ~ & \multicolumn{3}{c}{English Data} & \multicolumn{3}{c}{French Data} & \multicolumn{3}{c}{Spanish Data} & \multicolumn{3}{c}{German Data} & ~\\
  \textbf{Entity Class} & \textbf{Train} & \textbf{Dev} & \textbf{Test} & \textbf{Train} & \textbf{Dev} & \textbf{Test} & \textbf{Train} & \textbf{Dev} & \textbf{Test} & \textbf{Train} & \textbf{Dev} & \textbf{Test} & \textbf{Total}\\
  \midrule
  PER. & 3,340 & 419 & 431 & 3,237 & 396 & 362 & 3,186 & 393 & 362 & 3,213 & 395 & 380 & 16,114\\
  LOC. & 4,605 & 541 & 504 & 2,984 & 376 & 310 & 2,023 & 254 & 267 & 2,560 & 350 & 345 & 15,119\\
  ORG. & 7,342 & 927 & 968 & 1,511 & 204 & 183 & 2,115 & 277 & 247 & 1,556 & 181 & 205 & 15,716\\
  MISC. & 23,621 & 2,920 & 3,013 & 3,331 & 438 & 390 & 3,184 & 396 & 430 & 3,392 & 530 & 425 & 42,070\\
  \midrule
  Total & 38,908 & 4,807 & 4,916 & 11,063 & 1,414 & 1,245 & 10,508 & 1,320 & 1,306 & 10,721 & 1,456 & 1,355 & 89,019\\
  \midrule
  \textbf{Sent Num} & 19,966 & 2,463 & 2,533 & 4,812 & 609 & 561 & 4,798 & 591 & 593 & 4,750 & 628 & 604 & 42,908\\
  \bottomrule
\end{tabularx}
\end{table*}

\subsection{Dataset Comparison}\label{sec:dataset_comparison}

In Table \ref{Dataset_comparison}, we conduct a comparison bwtween our MMNER dataset and other representative NER datasets. On the one hand, multilingual NER datasets, including CoNLL2002 \cite{DBLP:conf/conll/Sang02}, CoNLL2003 \cite{DBLP:conf/conll/SangM03}, and WikiAnn \cite{DBLP:conf/acl/PanZMNKJ17}, are statisticed and compared with our dataset. CoNLL2002 and CoNLL2003 contain two languages each, while WikiAnn includes four languages. On the other hand, we compare our dataset with three widely used multimodal NER datasets: Twitter-2015 \cite{DBLP:conf/aaai/0001FLH18}, Twitter-2017 \cite{DBLP:conf/acl/JiZCLN18}, and CNERTA \cite{DBLP:conf/acl/SuiT00020}. Twitter-2015 and Twitter-2017 are text-image corpora, while CNERTA is a text-speech corpus.

Compared with all other existing datasets, we discover that our MMNER dataset is a large-scale corpus with the characters of two modalities and four languages. According to our investigation, this is the first publicly available dataset specifically designed for the multilingual and multimodal NER task.

\begin{table*}[htbp]
\caption{Dataset comparison bwtween our MMNER dataset and other NER datasets. AR, DE, DL, HI, EN, ES, ZH are the abbreviations of Arabic, German, Dutch, Hindi, English, Spanish and Chinese.}
\label{Dataset_comparison}
\setlength{\tabcolsep}{0.27cm}
\begin{tabular*}{\textwidth}{@{} lllrrrr@{} }
    \toprule
    Dataset & Multilingual & Multimodal & Train & Dev & Test & Total\\
    \midrule
    CoNLL2002 \cite{DBLP:conf/conll/Sang02} & \ding{51} (ES, DL) & \ding{55} (Text) & 24,129 & 4,810 & 6,712 & 35,651\\
    CoNLL2003 \cite{DBLP:conf/conll/SangM03} & \ding{51} (EN, DE) & \ding{55} (Text) & 27,692 & 6,534 & 6,844 & 41,070\\
    WikiAnn \cite{DBLP:conf/acl/PanZMNKJ17} & \ding{51} (EN, AR, HI, ZH) & \ding{55} (Text) & 65,000 & 31,000 & 31,000 & 127,000\\
    \midrule
    Twitter-2015 \cite{DBLP:conf/aaai/0001FLH18} & \ding{55} (EN) & \ding{51} (Text, Image) & 4,000 & 1,000 & 3,257 & 8,257\\
    Twitter-2017 \cite{DBLP:conf/acl/JiZCLN18} & \ding{55} (EN) & \ding{51} (Text, Image) & 3,373 & 723 & 723 & 4,819\\
    CNERTA \cite{DBLP:conf/acl/SuiT00020} & \ding{55} (ZH) & \ding{51} (Text, Speech) & 34,102 & 4,440 & 4,445 & 42,987\\
    \midrule
    \dataset{} (Ours) & \ding{51} (EN, FR, ES, DE) & \ding{51} (Text, Image) & 34,326 & 4,291 & 4,291 & 42,908\\
    \bottomrule
\end{tabular*}
\end{table*}

\section{The Proposed Method}\label{sec:methodology}

To address the NER challenge through the utilization of multimodal and multilingual information, we introduce a new MMNER task and design an innovative framework that validates the amalgamation of these different types of information. As for this part, we firstly provide a summary of the proposed model and a task definition for our MMNER task. Then, we utilize a feature extraction module to obtain representations of both text and images, which are aligned using multimodal alignment module. Additionally, the mechanism of multimodal interaction is described in detail, followed by a conditional random field (CRF) decoder to predict the named entities.

\textbf{Task Definition.} 
In our multilingual and multimodal NER (MMNER) task, we use a NER dataset that includes multilingual text in four languages: English, French, Spanish, and German. Each sentence $s=(t_{1}, t_{2}, \ldots, t_{n})$ in the corpus has a corresponding image $i_{s}$, and each token in the sentence $s$ is associated with a predefined label $y_{l}\left(1 \leq l \leq n\right)$. MMNER aims to assign a predefined label $y\left(y \in \mathcal{Y}\right)$ to every token in a sentence, and $\mathcal{Y}$ refers to the collection of all predefined labels. There are four categories of named entities (person, location, organization and miscellaneous) in this task and we adopt IOB2 tagging scheme for entity annotation. Note that if a sentence has multiple images, it can be converted into multiple sentence-image pairs.

\subsection{Overview of the Proposed Model}

The general framework of our proposed model is presented in Fig. \ref{architecture}. To begin with, we employ a multilingual BERT \cite{DBLP:conf/naacl/DevlinCLT19} encoder to acquire contextualized representation for the input sentence. The corresponding image of the sentence is embedded using a ViT \cite{DBLP:conf/iclr/DosovitskiyB0WZ21} encoder and a ResNet \cite{DBLP:journals/isci/OskoueiBM23} encoder to capture patch features and convolution features, respectively. To align the text and visual embeddings in the same subspace, we incorporate a multimodal alignment module that operates on top of these encoders. Meanwhile, the text representation and image representation interact with each other through two multimodal collaboration modules, which are based on the attention mechanism. Finally, at the top of the model, we include a CRF layer to predict entities with specific predefined labels.

\begin{figure*}[htbp]
\centering
\includegraphics[scale=0.365]{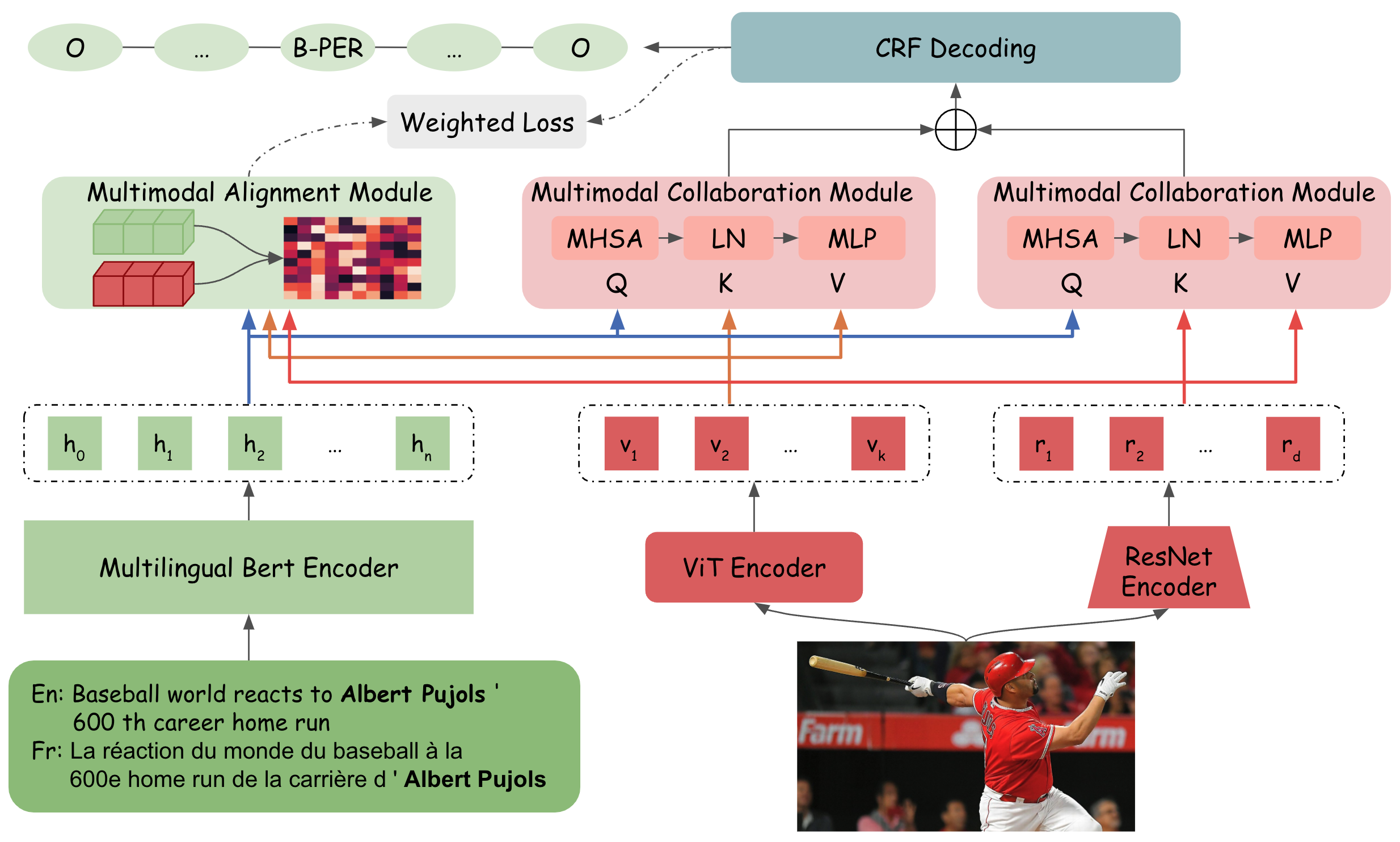}
\caption{The overall architecture of \model{}.}
\label{architecture}
\end{figure*}

\subsection{Feature Extraction Module}

To illustrate the process of feature acquisition with our input data, the details of the feature extraction module are specified in this subsection, including multilingual text representation and image representation.

\subsubsection{Multilingual Text Representation}

As depicted in Fig. \ref{architecture}, a Transformer \cite{DBLP:journals/apin/SunZTXW24} encoder is employed to learn text representations. Specifically, we use the first $M$ layers of a BERT multilingual base model\footnote{https://huggingface.co/bert-base-multilingual-cased}, which has been pretrained on 104 languages using a masked language model objective. Each sentence $s=(t_{1}, t_{2}, \ldots, t_{n})$ adds special tokens [CLS] and [SEP] to suit the model. $s^{\prime}=(t_{0}, t_{1}, t_{2}, \ldots, t_{n}, t_{n+1})$ is the processed sentence in which $t_{0}$ represents [CLS] and $t_{n+1}$ represents [SEP]. Besides, as calculated in Eq. (\ref{eq_t_1}) position embeddings $s_{pos}$ are added to provide position information for all tokens. 
\begin{equation}
s_{0} = s_{e} + s_{pos}
\label{eq_t_1}
\end{equation}

\noindent where $s_{e} \in \mathbb{R}^{(n+2) \times d}$ represents the token embedding matrix. To obtain contextualized representations, $M$ layers of multi-headed self-attention (MHSA) and multilayer perceptron (MLP) followed by layernorm (LN) in Transformer are adopted to calculate token embedding $s_{m}$ ($1 \leq m \leq M$) iteratively. The whole process is calculated by Eq. (\ref{eq_t_2}) and Eq. (\ref{eq_t_3}).
\begin{equation}
\tilde{s}_{m} = LN(MHSA(s_{m-1})) + s_{m-1} 
\label{eq_t_2}
\end{equation}\vspace{-0.5cm}
\begin{equation}
s_{m} = LN(MLP(\tilde{s}_{m})) + \tilde{s}_{m} \qquad 1 \leq m \leq M
\label{eq_t_3}
\end{equation}

\subsubsection{Image Representation}

Based on extensive research, ViT is capable of capturing both patch features and salient features without reducing image resolution or sacrificing local information. ResNet, on the other hand, learns convolutional features through its convolutional structure and has the ability to capture the general structure of image data, making it widely applicable. To enhance the extraction of image features across various dimensions, our model utilizes a ViT encoder to capture patch features and a ResNet encoder to extract convolution features. First, similar to the text representation process described above, a Transformer encoder is used to obtain image representations. Concretely, the first $K$ layers of ViT are used to encode the images, allowing the model to find out some salient features of the image at multiple levels of abstraction. For adapting the standard Transformer, a reshaped 2D image $x \in \mathbb{R}^{C \times H \times W}$ is split into $N = HW / P^{2}$ flattened 2D patches, where $C$ refers to the quantity of channels, $H$ and $W$ stand for the height and width of the image, $P^{2}$ is the resolution of every patch. Then, based on Eq. (\ref{eq1}), the flattened 2D patches are projected to $D_{v}$ dimensions. Note that absolute position embeddings $\mathbf{V}_{pos}$ are included to enable our image encoder to obtain position information of all patches. This enables the model to capture the spatial relationships between different patches in the image and ensure that the relative position of each patch is preserved during the encoding process.
\begin{equation}
\begin{split}
v_{0} = [x_{patch}^{1}\mathbf{V};x_{patch}^{2}\mathbf{V};\ldots;x_{patch}^{N}\mathbf{V}] + \mathbf{V}_{pos} \\
\mathbf{V} \in \mathbb{R}^{(C \times P^{2}) \times D_{v}}, \mathbf{V}_{pos} \in \mathbb{R}^{N \times D_{v}}
\end{split}
\label{eq1}
\end{equation}

Finally, as calculated in Eq. (\ref{eq2}) and Eq. (\ref{eq3}), $L$ layers of multi-headed self-attention (MHSA) and multilayer perceptron (MLP) in the Transformer model are applied to the image embedding $v_{k}$ ($1 \leq k \leq K$). Besides, layernorm (LN) and residual connections are also applied in every layer of the ViT model.
\begin{equation}
\tilde{v}_{k} = MHSA(LN(v_{k-1})) + v_{k-1} 
\label{eq2}
\end{equation}\vspace{-0.5cm}
\begin{equation}
v_{k} = MLP(LN(\tilde{v}_{k})) + \tilde{v}_{k} \qquad 1 \leq k \leq K
\label{eq3}
\end{equation}

As indicated in the lower right corner of Fig. \ref{architecture}, we utilize a ResNet encoder with deep layers to extract convolution features. To employ a pre-trained 152 layers ResNet, each image is transformed to a fixed size with 224 $\times$ 224 pixels. The encoder splits each image into multiple 7 $\times$ 7 small blocks and transforms each block into a 2048 dimensional vector. Formally, let $\mathbf{R}$ represent the overall representation of an image. It is mapped to the text representation space by using the formula $\mathbf{R^{\prime}} = \mathbf{W}_r^{\top} \mathbf{R} + \mathbf{b}$, where $\mathbf{W}_r \in \mathbb{R}^{2048 \times d}$ is the transition matrix and $\mathbf{b}$ is the bias.

\subsection{Multimodal Alignment Module}

Most previous studies learn text and image representations separately, which poses challenges for the interaction between the text and image encoders due to their distinct subspaces. To alleviate this problem, we align these two types of representations through contrastive learning (CL). Drawing inspiration from NT-Xent \cite{DBLP:conf/icml/ChenK0H20} and MoCo \cite{DBLP:conf/cvpr/He0WXG20}, our contrastive loss is transformed from visual representation learning to multimodal (text and image) representation learning. Specifically, the text and image representations in a batch are indicated as $h_{t}$ and $h_{i}$ respectively. To bring them into alignment, we initially resize them to the same dimensions using Eq. (\ref{eq8}) and Eq. (\ref{eq9}). Here, the hidden size we utilize is 768, which is the same as the text hidden size.
\begin{equation}
h_{t}^{\prime} = MLP(ReLU(MLP(h_{t})))
\label{eq8}
\end{equation}\vspace{-0.5cm}
\begin{equation}
h_{i}^{\prime} = MLP(ReLU(MLP(h_{i})))
\label{eq9}
\end{equation}

Different from contrastive learning that aligns two augmented views of the same image in visual models, the text representation $h_{t}^{\prime}$ and image representation $h_{i}^{\prime}$ are aligned by contrastive loss in our model. Specifically, assume that there are $N$ text-image pairs in one batch, and we can get $2N$ data pairs. Each data pair contains a text embedding and an image embedding. Only the original text-image pair and the image-text pair are positive, while the other $2N-2$ pairs are treated as negative examples. Hence, the contrastive loss can be formulated as follows:
\begin{equation}
l_{h_{t}^{\prime}, h_{i}^{\prime}} = \frac{ exp(sim(h_{t}^{\prime}, h_{i}^{\prime}) / \tau) }{ \sum_{t=1}^{N} \sum_{i=1}^{N} \mathbb{1}_{[t \neq i]} exp(sim(h_{t}^{\prime}, h_{i}^{\prime}) / \tau) }
\label{eq10}
\end{equation}

\begin{equation}
\mathcal{L}_{CL} = \frac{1}{2N} \sum_{t=1}^{N} \sum_{i=1}^{N} l_{h_{t}^{\prime}, h_{i}^{\prime}}
\label{eq11}
\end{equation}

In Eq. (\ref{eq10}), $sim(h_{t}^{\prime}, h_{i}^{\prime})$ represents the cosine similarity between the text vector representation and its image vector representation, $\tau$ is a temperature parameter, and $\mathbb{1}_{[t \neq i]}$ represents an indicator function whose value is 1 iff $t \neq i$, otherwise it is 0. The total contrastive loss for all text-image pairs is defined in Eq. (\ref{eq11}).

\subsection{Multimodal Collaboration Module}

To create a multimodal interaction between text and image, we employ a crossmodal attention layer to fuse multimodal information. As shown in Fig. \ref{architecture}, two multimodal collaboration modules are connected to the ViT encoder and the ResNet encoder respectively. Each module consists of a multi-headed self-attention layer and a multilayer perceptron. Besides, a residual connection and a layer normalization are used subsequent to each sub-layer. To be specific, we use the token embedding $T = (t_{1}, t_{2}, \ldots, t_{n})$ calculated by the text encoder to represent the query. It is referred to as $s_{M}$ in Section 4.2.1. Meanwhile, we use the image representation $V \in \mathbb{R}^{v \times d}$ obtained from the image encoder as the key-value pairs. It is referred to as $v_{K}$ or $\mathbf{R^{\prime}}$ in Section 4.2.2. In Fig. \ref{architecture}, ``Q", ``K", and ``V" represent the query, key, and value respectively. Formally,
\begin{equation}
IA_{i} = softmax(\frac{ \left[W_{q_{i}} T\right]^{\top} \left[W_{k_{i}} V\right] }{ \sqrt{d / m} }) \left[W_{v_{i}} V\right]^{\top}
\end{equation}

\begin{equation}
IA = W^{\prime} \left[ IA_{1}; IA_{2}; \ldots; IA_{m} \right]^{\top}
\label{eq_IA}
\end{equation}

\noindent where $IA_{i}$ denotes the $i$-th attention head between text and image, $\left \{ W_{q_{i}}, W_{k_{i}}, W_{v_{i}} \in \mathbb{R}^{d / m \times d} \right \}$ refer to all of the learnable parameters for each query, key, and value, and $W^{\prime} \in \mathbb{R}^{d \times d}$ in Eq. (\ref{eq_IA}) stands for the weight matrix for multi-head attention. The multimodal representation $IA \in \mathbb{R}^{d \times n}$ is calculated by the multi-headed self-attention layer. Afterwards, we calculate the final multimodal representation $R = (r_{1}, r_{2}, \ldots, r_{n})$ using the following formulas:
\begin{equation}
\widetilde{T} = LN (IA + T)
\end{equation}\vspace{-0.0cm}
\begin{equation}
\boldsymbol{R} = LN(\widetilde{T} + MLP(\widetilde{T}))
\end{equation}

\subsection{CRF Decoder}

Once we have fused the multimodal information, we obtain two multimodal representations from the ViT encoder and the ResNet encoder, respectively. These two final multimodal representations are concatenated together, and then fed into a standard CRF decoder to perform the MMNER task. The decoder predicts labels for each sentence $s$ and its corresponding image $v$. The probability of the label result $y$ is calculated using the following formulas:
\begin{equation}
score(\boldsymbol{y}, \boldsymbol{R}) = \sum_{i=1}^{n} E_{r_i, y_i} + \sum_{i=1}^n T_{y_i, y_{i+1}}
\end{equation}
\begin{equation}
P(\boldsymbol{y} | S, I) = \frac{ exp(score(\boldsymbol{y}, \boldsymbol{R})) }{ \sum_{\boldsymbol{\hat{y}}} exp(score(\boldsymbol{\hat{y}}, \boldsymbol{R})) }
\label{eq21}
\end{equation}

\noindent where $E_{i, y_i}$ represents the emission value of tag $y_i$ for the $i$-th multimodal representation of $R$, $T_{y_i, y_{i+1}}$ represents the transition value from tag $y_i$ to tag $y_{i+1}$. The conditional probability of the tag result $\boldsymbol{y}$ is derived by a normalization with the sum of all emission and transition scores over all possible label sequences, as defined in Eq. (\ref{eq21}). At the training stage, a negative log-likelihood loss is used to maximize the probability of the true label sequence, and the likelihood is formulated in Eq. (\ref{log_likelihood}):
\begin{equation}
\mathcal{L}_{CRF} = - log(P(\boldsymbol{\bar{y}} | S, I))
\label{log_likelihood}
\end{equation}

\noindent where $\bar{y}$ is the true label sequence of a sentence, and the conditional probability is calculated by Eq. (\ref{eq21}).

\subsection{Model Training}

According to the overall architecture depicted in Fig. \ref{architecture}, our MMNER model has one multimodal alignment module and a CRF decoder which generate losses. Overall, we aim to train \model{} via a weighted summation of two contrastive losses and one negative log-likelihood loss as follows:
\begin{equation}
\mathcal{L} = \alpha \mathcal{L}_{CRF} + (1-\alpha)(\mathcal{L}_{CL}^{\prime} + \mathcal{L}_{CL}^{\prime\prime})
\label{total_loss}
\end{equation}

\noindent where $\mathcal{L}_{CL}^{\prime}$ and $\mathcal{L}_{CL}^{\prime\prime}$ are the contrastive losses calculated by Eq. (\ref{eq11}), corresponding to ViT and ResNet respectively. The hyperparameter $\alpha$ is used to control the loss ratio.

\section{Experiments}\label{sec:experiments}

In our experiments, we select some comparative and representative baselines and compare our model with them on \dataset{}. Based on various experimental settings, we have done numerous multilingual and multimodal experiments. 
The following research questions (RQs) guide our experiments. We firstly introduce the implementation details and the compared baselines, then our experiments are conducted through these RQs.

\begin{itemize}
  \item \textbf{RQ1:} How does \model{} model compared with existing NER models?
  \item \textbf{RQ2:} What is the impact of multilingualism?
  \item \textbf{RQ3:} What is the impact of multimodality?
  \item \textbf{RQ4:} How challenging is the data from different sources?
  \item \textbf{RQ5:} How consistent is the annotation of \dataset{}?
\end{itemize}

\subsection{Implementation Details}

To efficiently verify our model \model{}, we conduct all the experiments on an Ubuntu server with PyTorch 1.12.1 and a NVIDIA GeForce RTX 3090 GPU. For the comparison methods, we adopt the original or reproduced public code, and the hyperparameters are consistent with the original papers. To adapt to multilingual scenarios, these comparison methods utilize GloVe\footnote{https://nlp.johnsnowlabs.com/2020/01/22/glove\_6B\_300.html} or multilingual BERT for encoding the multilingual text.

In our \model{} framework, we use a BERT multilingual base model with 12 transformer blocks for multilingual text representation. Besides, a pre-trained 12-layer CLIP \cite{DBLP:conf/icml/RadfordKHRGASAM21} and a pre-trained 152-layer ResNet\footnote{https://download.pytorch.org/models/resnet152-b121ed2d.pth} are used as image feature extractors. Both the text encoder and the image feature extractors are fine-tuned during the training phase. Specifically, some important hyperparameters of \model{} are presented in Table \ref{hyper_parameters}. 
The loss weight $\alpha$ in Eq. (\ref{total_loss}) and the learning rate are set to 0.8 and 5e-5 respectively, which obatin the best results on our validation dataset through a grid search across various combinations of [0.1, 0.9] and [1e-5, 5e-4]. Like the dropout in the BERT model, the parameters of all dropout layers in \model{} are set to 0.1. To improve the model's performance, we adopt a linear decay scheme that decays the learning rate from a fixed value to zero by the end of the training epochs.

\begin{table*}[htbp]
\caption{Some important hyperparameters of our model}
\label{hyper_parameters}
\begin{tabular*}{\textwidth}{@{} llll@{} }
  \toprule
  Hyperparam. & Value & Hyperparam. & Value \\
  \midrule
  $\alpha$ & 0.8 & Learning rate & 5e-5\\
  Batch size & 16 & Dropout & 0.1\\
  Learning scheduler & linear & Warmup steps & 0\\
  Text hidden size & 768 & Image hidden size & 768\\
  ViT encoder & clip-vit-base-patch32\tablefootnote{https://huggingface.co/openai/clip-vit-base-patch32} & Layers of multilingual BERT & 12\\
  Resolution of each image in ViT & 224 & Resolution of each image in ResNet & 224\\
  Optimizer & Adam & & \\
  \bottomrule
\end{tabular*}
\end{table*}

\subsection{Comparison Models}\label{sec:comp_methods}

To demonstrate the better performance of our approach, we first select some representative text-based NER methods for comparison (the first four models). Text-based NER models only consider text modal information, which lacks modal diversity. Therefore, several existing multimodal NER models incorporate visual information to overcome the above barrier. Here, we also pick out several classical and competitive multimodal models to illustrate the superiority of \model{} (the last four models).

\textbf{BiLSTM-CRF.}
Huang et al. \cite{DBLP:journals/corr/HuangXY15} provided an architecture of bidirectional LSTM combined with a CRF module for the sequence tagging task. This model automatically extracts past and future features, which gets rid of heavy dependence on hand-crafted features.

\textbf{HBiLSTM-CRF.}
Numerous studies show that character features can provide more morphological information. Lample et al. \cite{DBLP:conf/naacl/LampleBSKD16} improved the BiLSTM-CRF model by combining word embeddings and character-level word representations, which are obtained by an LSTM layer.

\textbf{BERT.}
By using a deep bidirectional Transformer \cite{DBLP:journals/apin/SunZTXW24}, Devlin et al. \cite{DBLP:conf/naacl/DevlinCLT19} designed a novel language model called BERT, which learns contextual vector representations for each word. When it is used for the NER task, BERT adds a softmax layer to predict labels based on its contextual embeddings.

\textbf{BERT-CRF.}
The BERT-CRF model is a variant of \textbf{BERT}, which stacks a multi-layer bidirectional Transformer at the bottom and a CRF decoder at the top.

\textbf{AdaCAN.}
To leverage textual and visual information, Zhang et al. \cite{DBLP:conf/aaai/0001FLH18} presented an adaptive co-attention network to obtain token-aware image representations and image-aware token representations through token-based visual attention and image-based textual attention, respectively. This work is based on \textbf{CNN-BiLSTM-CRF} \cite{DBLP:conf/acl/MaH16}, which utilizes a CNN layer to learn character-level word representations.

\textbf{UMT}
To tackle the issue of word representation's sensitivity to visual context, Yu et al. \cite{DBLP:conf/acl/YuJYX20} designed a multimodal interaction module to generate word-related visual representations and image-related word representations. They also utilized an entity span detection task to help predict entities with a unified multimodal transformer.

\textbf{UMGF}
Zhang et al. \cite{DBLP:conf/aaai/ZhangWLWZZ21} introduced a multimodal graph to build different types of semantic associations between words and visual objects. Then, they designed a multimodal fusion module to learn node representations, and finally predicted entity annotation using the attention-based multimodal representations with a CRF layer.

\textbf{MKGFormer}
To build a unified model architecture for various entity and relation extraction tasks like multimodal entity extraction and link prediction, Chen et al. \cite{DBLP:conf/sigir/ChenZLDTXHSC22} utilized a unified input-output in a hybrid transformer architecture. The architecture contains coarse-grained prefix-guided interaction to mitigate modal heterogeneity and fine-grained correlation-aware integration to reduce error sensitivity.

\subsection{Metrics}

The most commonly used evaluation metrics for NER tasks are precision, recall, and F1. Precision measures the accuracy of positive prediction made by a model. Recall represents the ratio of correctly predicted positive instances to the total number of actual positive instances. The F1 score is a combined metric that balances both precision and recall. Formulas \ref{precision}, \ref{recall}, and \ref{F1} provide the calculation methods for precision, recall, and F1 score respectively. $TP$ represents the number of entities correctly predicted as positive, $FP$ denotes the number of entities incorrectly predicted as positive, and $FN$ represents the number of entities incorrectly predicted as negative.

\begin{equation}
precision = \frac{TP}{TP + FP}
\label{precision}
\end{equation}

\begin{equation}
recall = \frac{TP}{TP + FN}
\label{recall}
\end{equation}

\begin{equation}
F1 = \frac{2 * precision}{precision + recall}
\label{F1}
\end{equation}

\subsection{Results}
Based on the research questions \textbf{RQ1:} to \textbf{RQ5:}, we conducted numerous experiments, and the following are the experimental results.

\subsubsection{RQ1: Overall Comparison with Existing Models}

We compare \model{} model with all the compared baselines in the overall dataset and each individual language of it. Table \ref{result_all} represents the F1 values for each individual type, as well as the precision, recall and F1 values across five different datasets. The maximum value in each column of the result tables is indicated in bold. By analyzing these large numbers of experiments, we obtain the following findings:

First, comparing all the text-based NER methods, the table shows that BERT-based models have better performance than LSTM-based models across all datasets. This indicates that the fine-tuning mechanism of the pre-trained model has a huge advantage in the NER task. Moreover, the combination of BERT and CRF can improve the F1 score except the German data on which drops a little, indicating that CRF can establish good constraints for tag sequence.

Second, when comparing four classical and competitive multimodal approaches with the above text-based NER models, we can observe that AdaCAN has the lowest overall F1 score and UMT has the highest overall F1 score across all datasets. Therefore, multimodal approaches are not always better than unimodal methods, and how to properly embed images within the model is a significant challenge for researchers.

Third, our proposed architecture (\model{}) obtains the best results in comparison with all the text-based NER models and all other multimodal NER approaches. We judge that the convolution features provided by our ResNet module and the patch features from the ViT encoder provides a rich vein of helpful information, which is aligned with corresponding text by contrastive loss. Besides, our multimodal collaboration module fuses this helpful information via multi-head attention to boost the performance.

\begin{table*}[htbp]
\caption{Experiments in all languages and four single languages on \dataset{}}
\label{result_all}
\setlength{\tabcolsep}{0.124cm}
\begin{tabularx}{\textwidth}{@{} llllllllll@{} }
  \toprule
  ~ & ~ & ~ & \multicolumn{4}{c}{Single Type (F1)(\%)} & \multicolumn{3}{c}{Overall(\%)} \\
  Language & Methods & Modality & PER. & LOC. & ORG. & MISC. & Precision & Recall & F1 \\
  \midrule
  \multirow{9}{*}{All} & BiLSTM-CRF \cite{DBLP:journals/corr/HuangXY15} & Text & 91.12 & 55.79 & 55.52 & 56.32 & 65.07 & 59.57 & 62.20\\
  ~ & HBiLSTM-CRF \cite{DBLP:conf/naacl/LampleBSKD16} & Text & 91.76 & 57.64 & 58.98 & 58.57 & 67.24 & 61.71 & 64.36\\
  ~ & BERT \cite{DBLP:conf/naacl/DevlinCLT19} & Text & 94.48 & 63.36 & 63.96 & 62.17 & 66.54 & 69.98 & 68.22\\
  ~ & BERT-CRF & Text & 94.87 & 64.25 & 66.08 & 63.24 & 69.04 & 69.48 & 69.26\\
  \cmidrule(){2-10}
  ~ & AdaCAN \cite{DBLP:conf/aaai/0001FLH18} & Text + Image & 89.24 & 52.13 & 53.86 & 53.62 & 63.50 & 56.30 & 59.69\\
  ~ & UMT \cite{DBLP:conf/acl/YuJYX20} & Text + Image & 95.50 & 63.82 & 65.97 & \textbf{65.27} & 69.27 & 71.06 & 70.15\\
  ~ & UMGF \cite{DBLP:conf/aaai/ZhangWLWZZ21} & Text + Image & 93.49 & 62.11 & 63.61 & 62.20 & 67.52 & 68.12 & 67.82\\
  ~ & MKGFormer \cite{DBLP:conf/sigir/ChenZLDTXHSC22} & Text + Image & 94.58 & 60.03 & 61.15 & 59.42 & 64.96 & 66.47 & 65.71\\
  ~ & 2M-NER (Ours) & Text + Image & \textbf{96.34} & \textbf{65.00} & \textbf{67.14} & 64.75 & \textbf{69.64} & \textbf{71.35} & \textbf{70.49}\\
  \midrule
  \multirow{9}{*}{English} & BiLSTM-CRF \cite{DBLP:journals/corr/HuangXY15} & Text & 86.23 & 50.15 & 56.63 & 64.18 & 64.73 & 60.76 & 62.68\\
  ~ & HBiLSTM-CRF \cite{DBLP:conf/naacl/LampleBSKD16} & Text & 88.86 & 47.25 & 54.12 & 65.94 & 64.78 & 63.87 & 64.32\\
  ~ & BERT \cite{DBLP:conf/naacl/DevlinCLT19} & Text & 92.73 & 54.70 & 63.94 & 68.78 & 66.15 & 70.46 & 68.24\\
  ~ & BERT-CRF & Text & 92.94 & 55.74 & 64.36 & 70.46 & 68.77 & 70.53 & 69.64\\
  \cmidrule(){2-10}
  ~ & AdaCAN \cite{DBLP:conf/aaai/0001FLH18} & Text + Image & 85.21 & 46.04 & 50.44 & 62.12 & 61.87 & 58.76 & 60.28\\
  ~ & UMT \cite{DBLP:conf/acl/YuJYX20} & Text + Image & 93.53 & 55.79 & 62.30 & \textbf{72.60} & 69.46 & \textbf{71.59} & 70.51\\
  ~ & UMGF \cite{DBLP:conf/aaai/ZhangWLWZZ21} & Text + Image & 88.69 & 53.90 & 60.87 & 68.16 & 65.64 & 68.79 & 67.18\\
  ~ & MKGFormer \cite{DBLP:conf/sigir/ChenZLDTXHSC22} & Text + Image & 88.45 & 53.10 & 55.92 & 67.00 & 63.79 & 66.53 & 65.13\\
  ~ & 2M-NER (Ours) & Text + Image & \textbf{96.90} & \textbf{58.33} & \textbf{64.45} & 71.40 & \textbf{70.05} & 71.51 & \textbf{70.77}\\
  \midrule
  \multirow{9}{*}{French} & BiLSTM-CRF \cite{DBLP:journals/corr/HuangXY15} & Text & 88.39 & 65.13 & 54.67 & 25.57 & \textbf{71.55} & 53.14 & 60.98\\
  ~ & HBiLSTM-CRF \cite{DBLP:conf/naacl/LampleBSKD16} & Text & 89.32 & 65.64 & 54.48 & 34.10 & 68.92 & 57.59 & 62.74\\
  ~ & BERT \cite{DBLP:conf/naacl/DevlinCLT19} & Text & 90.96 & \textbf{67.74} & 62.79 & 35.08 & 59.59 & 66.44 & 62.83\\
  ~ & BERT-CRF & Text & 92.24 & 62.33 & 59.67 & \textbf{37.34} & 63.10 & 62.75 & 62.92\\
  \cmidrule(){2-10}
  ~ & AdaCAN \cite{DBLP:conf/aaai/0001FLH18} & Text + Image & 86.82 & 58.58 & 50.53 & 21.00 & 66.77 & 49.01 & 56.53\\
  ~ & UMT \cite{DBLP:conf/acl/YuJYX20} & Text + Image & 94.07 & 67.56 & 61.71 & \textbf{37.34} & 62.08 & 66.32 & 64.13\\
  ~ & UMGF \cite{DBLP:conf/aaai/ZhangWLWZZ21} & Text + Image & 91.38 & 65.58 & 55.53 & 29.40 & 58.80 & 62.40 & 60.54\\
  ~ & MKGFormer \cite{DBLP:conf/sigir/ChenZLDTXHSC22} & Text + Image & 89.32 & 61.52 & 49.86 & 27.45 & 56.73 & 58.29 & 57.50\\
  ~ & 2M-NER (Ours) & Text + Image & \textbf{95.34} & 67.04 & \textbf{63.39} & 36.10 & 62.24 & \textbf{67.43} & \textbf{64.73}\\
  \midrule
  \multirow{9}{*}{Spanish} & BiLSTM-CRF \cite{DBLP:journals/corr/HuangXY15} & Text & 83.73 & 62.28 & 69.60 & 46.15 & \textbf{72.29} & 59.54 & 65.30\\
  ~ & HBiLSTM-CRF \cite{DBLP:conf/naacl/LampleBSKD16} & Text & 84.60 & 66.54 & 68.10 & 50.20 & 71.64 & 63.05 & 67.07\\
  ~ & BERT \cite{DBLP:conf/naacl/DevlinCLT19} & Text & 89.93 & 70.61 & 68.64 & 52.67 & 68.83 & 71.35 & 70.07\\
  ~ & BERT-CRF & Text & 90.01 & 72.03 & 67.98 & 54.92 & 71.75 & 70.38 & 71.06\\
  \cmidrule(){2-10}
  ~ & AdaCAN \cite{DBLP:conf/aaai/0001FLH18} & Text + Image & 85.44 & 56.97 & 66.67 & 47.50 & 70.51 & 59.69 & 64.65\\
  ~ & UMT \cite{DBLP:conf/acl/YuJYX20} & Text + Image & 90.88 & 71.24 & \textbf{73.79} & \textbf{55.33} & 70.15 & 72.82 & 71.46\\
  ~ & UMGF \cite{DBLP:conf/aaai/ZhangWLWZZ21} & Text + Image & 88.40 & 68.85 & 70.18 & 50.30 & 66.26 & 71.09 & 68.59\\
  ~ & MKGFormer \cite{DBLP:conf/sigir/ChenZLDTXHSC22} & Text + Image & 88.52 & 60.47 & 69.73 & 47.52 & 64.39 & 66.87 & 65.61\\
  ~ & 2M-NER (Ours) & Text + Image & \textbf{93.08} & \textbf{73.40} & 70.70 & 55.12 & 71.33 & \textbf{73.13} & \textbf{72.22}\\
  \midrule
  \multirow{9}{*}{German} & BiLSTM-CRF \cite{DBLP:journals/corr/HuangXY15} & Text & 83.27 & 52.85 & 45.97 & 32.09 & \textbf{64.20} & 48.58 & 55.31\\
  ~ & HBiLSTM-CRF \cite{DBLP:conf/naacl/LampleBSKD16} & Text & 85.04 & 57.94 & 45.16 & 37.97 & 63.39 & 53.97 & 58.30\\
  ~ & BERT \cite{DBLP:conf/naacl/DevlinCLT19} & Text & 89.83 & 63.76 & 53.63 & 39.81 & 60.84 & 63.79 & 62.28\\
  ~ & BERT-CRF & Text & 88.69 & 61.89 & 55.85 & 39.32 & 63.55 & 60.31 & 61.88\\
  \cmidrule(){2-10}
  ~ & AdaCAN \cite{DBLP:conf/aaai/0001FLH18} & Text + Image & 82.37 & 49.13 & 47.09 & 32.57 & 57.34 & 50.33 & 53.61\\
  ~ & UMT \cite{DBLP:conf/acl/YuJYX20} & Text + Image & 89.47 & \textbf{64.02} & 52.76 & \textbf{44.99} & 62.37 & 64.09 & 63.22\\
  ~ & UMGF \cite{DBLP:conf/aaai/ZhangWLWZZ21} & Text + Image & 87.42 & 58.09 & 51.95 & 33.49 & 55.78 & 60.49 & 58.04\\
  ~ & MKGFormer \cite{DBLP:conf/sigir/ChenZLDTXHSC22} & Text + Image & 87.86 & 54.13 & 45.99 & 35.01 & 54.83 & 57.29 & 56.04\\
  ~ & 2M-NER (Ours) & Text + Image & \textbf{93.47} & 63.85 & \textbf{56.87} & 41.61 & 62.27 & \textbf{65.40} & \textbf{63.80}\\
  \bottomrule
\end{tabularx}
\end{table*}

\subsubsection{RQ2: The Effect of Multilingualism}\label{subsection_multilingualism_effect}

To study the effect of multilingualism, we add another language on the basis of a single language. For instance, English, French, and Spanish are combined with German respectively, and we use these datasets to recognize entities in any of the assembled languages. Table \ref{multilingual_effects} shows all the possible combinations and experimental results of our method on \dataset{}. From the table, we can find that: (1) In the four languages we use, English and German belong to the Germanic language family, while French and Spanish belong to the Romance languages. Moreover, adding English data on top of German data to test German, the F1 score increases by 1.34\%. The combination of French training data and Spanish training data can improve the F1 score by 1.76\% on the French test set and 1.62\% on the Spanish test set, respectively. Hence, although the F1 score calculated on our English test set does not improve much after adding German data to English data due to the relatively small scale of German data, we can still conclude that combining languages from the same language family can promote the MMNER task. (2) The combination of languages from different language families makes the MMNER task more challenging. For instance, adding the German training set to the Spanish training set does not improve the model's ability to predict Spanish test set. This is because the syntax and grammar of different language families will bring different degrees of confusion and interference to the model. (3) For German, French, and Spanish, increasing the training dataset of the same language can greatly improve the F1 scores of entity \textit{ORG} and entity \textit{MISC}. We consider that these two entity types are more sensitive to data size, so their corresponding F1 scores will increase rapidly when the amount of data increases.

\begin{table*}[htbp]
\caption{Experimental results of multilingual effects}
\label{multilingual_effects}
\setlength{\tabcolsep}{0.176cm}
\begin{tabular*}{\textwidth}{@{} lllllllll@{}}
  \toprule
  ~ & ~ & \multicolumn{4}{c}{Single Type (F1)(\%)} & \multicolumn{3}{c}{Overall(\%)} \\
  Language of training & Language of testing & PER. & LOC. & ORG. & MISC. & Precision & Recall & F1 \\
  \midrule
  English & English & 96.90 & 58.33 & 64.45 & 71.40 & 70.05 & 71.51 & 70.77\\
  English \& German & English & 95.16 & 55.61 & 64.26 & 71.93 & 70.38 & 71.18 & \textbf{70.78}\\
  English \& French & English & 94.34 & 57.38 & 65.07 & 70.69 & 69.41 & 70.99 & 70.19\\
  English \& Spanish & English & 93.50 & 57.50 & 63.98 & 71.61 & 69.16 & 71.82 & 70.47\\
  \midrule
  German & German &  93.47 & 63.85 & 56.87 & 41.61 & 62.27 & 65.40 & 63.80\\
  German \& English & German & 92.87 & 61.95 & 62.76 & 46.09 & 63.91 & 66.42 & \textbf{65.14}\\
  German \& French & German & 90.63 & 63.19 & 59.13 & 44.98 & 62.25 & 65.70 & 63.93\\
  German \& Spanish & German & 91.81 & 64.99 & 53.77 & 43.01 & 62.03 & 64.97 & 63.46\\
  \midrule
  French & French & 95.34 & 67.04 & 63.39 & 36.10 & 62.24 & 67.43 & 64.73\\
  French \& English & French & 93.04 & 65.53 & 61.28 & 36.96 & 61.54 & 66.08 & 63.73\\
  French \& German & French & 94.13 & 69.96 & 61.87 & 36.23 & 62.37 & 67.67 & 64.91\\
  French \& Spanish & French & 96.02 & 68.18 & 66.11 & 39.32 & 64.97 & 68.07 & \textbf{66.49}\\
  \midrule
  Spanish & Spanish & 93.08 & 73.40 & 70.70 & 55.12 & 71.33 & 73.13 & 72.22\\
  Spanish \& English & Spanish & 92.69 & 70.92 & 73.09 & 53.79 & 70.64 & 72.37 & 71.49\\
  Spanish \& German & Spanish & 92.54 & 69.84 & 72.48 & 55.95 & 70.09 & 73.51 & 71.76\\
  Spanish \& French & Spanish & 93.78 & 73.06 & 76.27 & 57.14 & 72.53 & 75.19 & \textbf{73.84}\\
  \bottomrule
\end{tabular*}
\end{table*}

\subsubsection{RQ3: The Effect of Multimodality}

For all the multimodal models including our model, we remove the image encoders from each model architecture to verify whether adding images can improve model performance. From Table \ref{result_multimodality}, we find that: (1) All the models except AdaCAN and UMGF have either major or minor improvement after adding images over single or multiple languages, which shows that the image features are indeed helpful to recognize entities, and our model achieves better F1 score across languages by adding images. (2) On the Spanish and German datasets, our model performs 0.16\% and 0.2\% lower in terms of the best F1 score compared to the baseline models, respectively. Across all language datasets, our model has a slight 0.02\% lower best F1 score compared to the baseline models. However, after incorporating multimodal features, our model consistently outperforms the baseline models in terms of F1 score, indicating that our model excels in the extraction and fusion of multimodal features. (3) For our model \model{}, incorporating multimodal features has improved the overall F1 score by 0.78\%, 1.6\%, 1.01\%, and 1.64\% for English, French, Spanish, and German, respectively. This indicates that the fusion of multimodal features is beneficial for entity recognition in each language, especially for French and German.

\begin{table*}[htbp]
\caption{Experimental results of different models without images}
\label{result_multimodality}
\setlength{\tabcolsep}{0.19cm}
\begin{tabular*}{\textwidth}{@{} llllllllll@{} }
  \toprule
  ~ & ~ & ~ & \multicolumn{4}{c}{Single Type (F1)(\%)} & \multicolumn{3}{c}{Overall(\%)} \\
  Language & Methods & Modality & PER. & LOC. & ORG. & MISC. & Precision & Recall & F1 \\
  \midrule
  \multirow{5}{*}{All} & AdaCAN \cite{DBLP:conf/aaai/0001FLH18} & Only Text & 88.18 & 51.79 & 53.14 & 52.73 & 63.39 & 55.31 & 59.07\\
  ~ & UMT \cite{DBLP:conf/acl/YuJYX20} & Only Text & 95.09 & 63.82 & 66.06 & 64.64 & 68.75 & 70.82 & \textbf{69.77}\\
  ~ & UMGF \cite{DBLP:conf/aaai/ZhangWLWZZ21} & Only Text & 93.64 & 62.04 & 62.44 & 60.89 & 66.03 & 67.85 & 66.93\\
  ~ & MKGFormer \cite{DBLP:conf/sigir/ChenZLDTXHSC22} & Only Text & 93.53 & 59.71 & 61.06 & 59.38 & 64.71 & 66.19 & 65.44\\
  ~ & 2M-NER (Ours) & Only Text & 94.53 & 63.73 & 65.33 & 64.86 & 68.89 & 70.62 & 69.75\\
  \midrule
  \multirow{5}{*}{English} & AdaCAN \cite{DBLP:conf/aaai/0001FLH18} & Only Text & 85.09 & 42.22 & 50.03 & 61.12 & 59.78 & 59.03 & 59.41\\
  ~ & UMT \cite{DBLP:conf/acl/YuJYX20} & Only Text & 92.52 & 56.05 & 63.91 & 71.36 & 68.83 & 71.11 & 69.95\\
  ~ & UMGF \cite{DBLP:conf/aaai/ZhangWLWZZ21} & Only Text & 88.29 & 55.31 & 61.60 & 68.26 & 65.24 & 69.55 & 67.32\\
  ~ & MKGFormer \cite{DBLP:conf/sigir/ChenZLDTXHSC22} & Only Text & 87.99 & 51.94 & 57.10 & 66.32 & 63.71 & 65.73 & 64.70\\
  ~ & 2M-NER (Ours) & Only Text & 96.74 & 53.61 & 63.47 & 71.27 & 68.85 & 71.16 & \textbf{69.99}\\
  \midrule
  \multirow{5}{*}{French} & AdaCAN \cite{DBLP:conf/aaai/0001FLH18} & Only Text & 86.31 & 61.05 & 52.85 & 23.42 & 62.54 & 52.10 & 56.85\\
  ~ & UMT \cite{DBLP:conf/acl/YuJYX20} & Only Text & 91.89 & 67.06 & 55.05 & 37.49 & 59.68 & 66.08 & 62.72\\
  ~ & UMGF \cite{DBLP:conf/aaai/ZhangWLWZZ21} & Only Text & 91.46 & 66.18 & 52.11 & 30.30 & 57.79 & 63.54 & 60.53\\
  ~ & MKGFormer \cite{DBLP:conf/sigir/ChenZLDTXHSC22} & Only Text & 88.65 & 61.61 & 52.17 & 26.14 & 56.46 & 57.96 & 57.20\\
  ~ & 2M-NER (Ours) & Only Text & 92.73 & 67.88 & 58.60 & 35.19 & 61.54 & 64.81 & \textbf{63.13}\\
  \midrule
  \multirow{5}{*}{Spanish} & AdaCAN \cite{DBLP:conf/aaai/0001FLH18} & Only Text & 83.88 & 56.97 & 68.64 & 45.78 & 68.31 & 59.39 & 63.54\\
  ~ & UMT \cite{DBLP:conf/acl/YuJYX20} & Only Text & 90.91 & 70.21 & 69.03 & 57.71 & 70.27 & 72.52 & \textbf{71.37}\\
  ~ & UMGF \cite{DBLP:conf/aaai/ZhangWLWZZ21} & Only Text & 88.73 & 69.15 & 68.35 & 50.85 & 65.40 & 71.78 & 68.44\\
  ~ & MKGFormer \cite{DBLP:conf/sigir/ChenZLDTXHSC22} & Only Text & 88.10  & 60.90 & 66.53 & 46.86 & 63.10 & 66.56 & 64.79\\
  ~ & 2M-NER (Ours) & Only Text & 92.32 & 71.22 & 70.89 & 54.91 & 68.85 & 73.74 & 71.21\\
  \midrule
  \multirow{5}{*}{German} & AdaCAN \cite{DBLP:conf/aaai/0001FLH18} & Only Text & 82.14 & 52.91 & 49.86 & 29.63 & 57.77 & 50.91 & 54.12\\
  ~ & UMT \cite{DBLP:conf/acl/YuJYX20} & Only Text & 89.27 & 64.56 & 50.74 & 44.03 & 61.52 & 63.22 & \textbf{62.36}\\
  ~ & UMGF \cite{DBLP:conf/aaai/ZhangWLWZZ21} & Only Text & 89.04 & 60.11 & 50.12 & 31.18 & 56.74 & 59.19 & 57.94\\
  ~ & MKGFormer \cite{DBLP:conf/sigir/ChenZLDTXHSC22} & Only Text & 87.02 & 53.97 & 44.92 & 33.33 &  53.98 & 56.25 & 55.09\\
  ~ & 2M-NER (Ours) & Only Text & 92.61 & 61.85 & 55.90 & 38.64 & 62.34 & 61.98 & 62.16\\
  \bottomrule
\end{tabular*}
\end{table*}

\subsubsection{RQ4: Challenging Analysis of Data from Different Sources}

In Section \ref{sec:dataset_construction}, we construct our dataset \dataset{} from mBART50 and Twitter-2017, which are two different datasets in some aspects. First, person entities are replaced with \emph{<PERSON>} token in mBART50 due to privacy, so all the person entities in \dataset{} are collected from Twitter-2017. Second, the data scale of mBART50 is larger than that of Twitter-2017, and most of the entities (LOC, ORG and MISC) come from mBART50. To evaluate the difficulty of different data sources, we analyse the F1 scores of four different entities from Table \ref{result_all}, Table \ref{multilingual_effects}, and Table \ref{result_multimodality}. The common feature of these tables is that the F1 score of \textit{PER} is much higher than other entities. For example, the F1 score of \textit{PER} is almost 30\% higher compared to the F1 value of \textit{LOC} in Table \ref{result_all}. The reason is that different datasets bring challenges of varying degrees of difficulty. On the one hand, as shown in Table \ref{data_examples}, an image in Twitter-2017 usually corresponds to a distinct person entity, so it is easy for this image to align with entities in sentences. On the other hand, an image in mBART50 usually corresponds to a whole sentence which has entities and numerous other words. This sentence-level alignment interferes a lot with the NER model, hence the MMNER task on mBART50 is extremely more challenging.

\subsubsection{RQ5: Model Validation for the Consistency of Dataset Annotation}\label{subsection_dataset_validation}

We use k-fold validation to measure the effectiveness of the \model{} model on the All Data described in Table \ref{MMNER_statistics}. Specifically, we divide the training dataset into five subsets of roughly equal size without any intersection and keep the development and test set as same as the main model. Then, we train the \model{} model ten times with image, every two times using a different subset as the training set. Table \ref{result_kfold} reports our average results, including the overall and individual type performance of \model{} across our different validation sets. From the table, those results indicate that the performance of each k-fold model converges and is reduced only a little compared to our ALL model, which indicates that the annotation of our dataset is consistent and efficient.

\begin{table*}[htbp]
\caption{K-fold experimental results on the all language dataset}
\label{result_kfold}
\setlength{\tabcolsep}{0.52cm}
\begin{tabular*}{\textwidth}{@{} llllllll}
 \toprule
  ~ & \multicolumn{4}{c}{Single Type (F1)(\%)} & \multicolumn{3}{c}{Overall(\%)} \\
 Data Type & PER. & LOC. & ORG. & MISC. & Precision & Recall & F1 \\
 \midrule
 K-fold-1 & 94.97 & 63.89 & 65.15 & 64.54 & 69.14 & 70.14 & 69.64 \\
 K-fold-2 &95.82 & 63.82 & 63.98 & 64.15 & 68.90 & 69.91 & 69.40 \\ 
 K-fold-3 & 94.75 & 62.61 & 63.38 & 64.05 & 68.33 & 69.38 & 68.85 \\
 K-fold-4 &94.91 & 62.50 & 63.67 & 63.49 & 68.26 & 69.01 & 68.63 \\ 
 K-fold-5 &95.02 & 63.15 & 63.21 & 63.31 & 68.41 & 68.82 & 68.62 \\
 \midrule
 ALL & 96.34 & 65.00 & 67.14 & 64.75 & 69.64 & 71.35 & \textbf{70.49} \\
\bottomrule
\end{tabular*}
\end{table*}

\subsection{Ablation Study}

In order to analyze the contributions of image features in our model, we conducted some ablation experiments. In Fig. \ref{architecture}, the model utilizes ResNet to learn convolutional features and leverages ViT to capture patch features. Both types of features interact with textual features through multimodal fusion. To understand the contributions of ResNet and ViT, we designed two sets of ablation experiments. One set removes only ResNet from the original model, while the other set removes only ViT. The experimental results are shown in Table \ref{ablation_study}.

\begin{table*}[htbp]
\caption{Ablation experiments on the feature contributions of ResNet and ViT}
\label{ablation_study}
\setlength{\tabcolsep}{0.52cm}
\begin{tabular*}{\textwidth}{@{} llllllll}
 \toprule
  ~ & \multicolumn{4}{c}{Single Type (F1)(\%)} & \multicolumn{3}{c}{Overall(\%)} \\
 Model & PER. & LOC. & ORG. & MISC. & Precision & Recall & F1 \\
 \midrule
 2M-NER & 96.34 & 65.00 & 67.14 & 64.75 & 69.64 & 71.35 & \textbf{70.49} \\
 -ResNet & 95.02 & 62.32 & 64.86 & 63.96 & 67.80 & 70.35 & 69.05 \\
 -ViT & 95.09 & 63.76 & 65.56 & 63.65 & 68.28 & 70.21 & 69.23 \\ 
\bottomrule
\end{tabular*}
\end{table*}

The ablation experiments were conducted on all language datasets, using the same data and parameters, with only the model architecture being different. From Table \ref{ablation_study}, it can be observed that removing either ResNet or ViT leads to a decrease in overall performance, while combining both yields better results. Additionally, removing ViT results in a smaller drop in F1 score compared to removing ResNet, indicating that ResNet contributes more significantly to the model's features.

\subsection{Error Analysis}

We observed two main types of errors in our model. One is related to the richness of image information, and the other is attributed to the diversity of entity semantics.

Regarding the image information, we have discovered that some errors occur when the images do not contain relevant entity information. Without the assistance of image-related cues, the model can only predict entity positions and types based on the textual context. For example, in the case of the sentence ``00:30 An Evening with Glen Campbell - Musicians play arrangements of Glen Campbell's hits at the Royal Festival Hall in 1977" on Twitter, the corresponding image shows a scene of musicians and a band performing in a dark background. Due to the inability to clearly see the music hall, the model incorrectly classifies the entity type of ``Royal Festival Hall" as MISC. Future work should focus on investigating how to enhance the scene knowledge and semantic information of images, in order to better determine the entity types mentioned in sentences.

Regarding entity semantics, some entities themselves can represent multiple meanings. Therefore, if the model lacks domain knowledge, it may misclassify the entity type. For example, in the sentence ``Jeep tour on the red rocks via Pink Adventure Tours," the word ``Jeep" can refer to both an American automotive brand and a type of off-road vehicle. Due to this ambiguity, the model may incorrectly predict the MISC type as ORG or vice versa. One future research direction to address this issue is to incorporate multidimensional domain knowledge to better understand the semantic context of sentences.

\section{Conclusion and Future Work}\label{sec:conclusion}

To sum up, we have constructed a large-scale and high-value MMNER dataset named \dataset{}, which is the first public dataset that supports both multilingual and multimodal. To facilitate research on \dataset{}, we design a general MMNER framework, named \model{}, that achieves the highest F1 score on this valuable dataset. Sepcifically, \model{} leverages contrastive learning to align language and vision representations, and utilizes a Transformer based multimodal collaboration module to effectively fuse multilingual texts and images. The results from our experimental tables show that \model{} achieves better performance than numerous baseline models. Further analysis indicates the positive effects of both multilingualism and multimodality.

In the future, we would like to conduct further studies in two directions. Firstly, as analyzed in our experiments, language combination of the same language family can promote the MMNER task, so we would like to test our model on more other languages. Secondly, according to our experimental results and previous works, more modalities can provide valuable information for the NER task. Therefore, we plan to introduce the acoustic modality, and propose an innovative framework that supports combined input of text, image, and sound.


\bmhead{Acknowledgements}
This work was supported by the National Key R\&D Program of China (No. 2023YFF0725600) and Major Special Funds of the Changsha Scientific and Technological Project (Grant No. kh2202006).

\bibliography{sn-bibliography}

\bibliographystyle{sn-vancouver}

\end{document}